\documentclass[11pt]{article}

\usepackage[margin=1in]{geometry}
\usepackage{graphicx}
\usepackage{booktabs}
\usepackage{amsmath,amssymb}
\usepackage{microtype}
\usepackage[hidelinks]{hyperref}
\usepackage[numbers,sort&compress]{natbib}
\usepackage{setspace}
\usepackage{lineno}

\title{Healthcare AI for Automation or Allocation? A Transaction Cost Economics Framework}
\author{Ari Ercole\\Department of Medicine, University of Cambridge, Addenbrooke's Hospital, \\Hills Road, Cambridge CB2 0QQ, UK}

\begin{document}
%\doublespacing
%\linenumbers

\maketitle

\begin{abstract}
Healthcare productivity is shaped not only by clinical complexity but by the costs of coordinating work under uncertainty. Transaction‑cost economics offers a theory of these coordination frictions, yet has rarely been operationalised at task level across health occupations. Using task statements and frequency weights from the O*NET occupational database, we characterised healthcare work at task granularity and coded each unique task using a constrained large language model into one dominant transaction‑cost category (information search, decision and bargaining, monitoring and enforcement, or adaptation and coordination) together with an overall transaction‑cost intensity score. Aggregating to the occupation level, clinician roles exhibited substantially higher transaction‑cost intensity than non‑clinician roles, driven primarily by greater burdens of information search and decision‑related coordination, while dispersion of transaction costs within occupations did not differ. These findings demonstrate systematic heterogeneity in the nature of coordination work across healthcare roles and suggest that the opportunities for digital and AI interventions are unevenly distributed, shaped less by technical task complexity than by underlying coordination structure.
\end{abstract}

\section{Introduction}

Digital tools and artificial intelligence (AI) are widely presented as a route to better outcomes, better staff experience, and improved productivity in healthcare. Yet translation from technical promise to operational impact remains uneven. Contemporary policy and professional guidance emphasise that the central determinant of benefit is not the existence of an AI model, but how it is implemented: whether it is tested in real clinical settings, integrated into workflows, governed safely, supported by adequate infrastructure, and aligned with incentives and accountability structures \citep{BMA2024AI}. The same guidance stresses that AI is not intrinsically beneficial: implemented well it can improve care, and implemented poorly it can worsen it \citep{BMA2024AI}. In parallel, implementation-focused editorials in \textit{npj Digital Medicine} have converged on the same point for generative AI: progress depends on evaluation frameworks, leadership and incentives for adoption, and continued regulation that addresses issues such as reliability and liability \citep{Raza2024Pathways}. These perspectives imply a practical question that remains unresolved for health systems: \emph{where, exactly, should AI sit within the work of healthcare?} 

One reason this question is hard is that healthcare work is not a single activity. It is a tightly coupled system of clinical judgement, information work, documentation, coordination, assurance, and adaptation under uncertainty. Many deployments of digital tools treat AI as an ``add-on'' to existing pathways, targeting visible downstream burdens such as documentation or administrative tasks \citep{BMA2024AI}. Some applications do demonstrate measurable productivity gains when the technology changes the effective division of labour within a pathway. For example, a cluster-randomized trial reported higher specialist clinic productivity when an autonomous AI performed a defined diagnostic function within a care pathway \citep{Abramoff2023Autonomous}. This illustrates a key point for digital strategy: productivity gains are possible, but depend on \emph{task allocation and pathway design}, not model capability alone.

We therefore require a framework that treats healthcare as an organisational system and asks which parts of work are structurally amenable to digital assistance. Transaction-cost economics (TCE) provides such a framework. In its simplest form, TCE distinguishes the cost of \emph{producing} a service from the cost of \emph{coordinating} the people and information required to deliver it. The classic insight is that many systems are constrained less by the technical difficulty of ``doing the thing'' and more by the overhead of making coordinated action possible \citep{Coase1937,Williamson1985,North1990}. In healthcare, these coordination frictions appear as: (i) \textbf{search and information} costs (finding relevant information and making it usable), (ii) \textbf{bargaining and decision} costs (reaching commitments across actors and services), (iii) \textbf{monitoring and enforcement} costs (documentation, assurance, audit and compliance), and (iv) \textbf{adaptation and coordination} costs (replanning, handovers, and exception handling when circumstances change) \citep{Coase1937,Williamson1985,North1990}. TCE is useful here not as economic theory for its own sake, but because it provides a vocabulary to connect digital tools to the \emph{specific} frictions they reduce (or sometimes increase).

This framing changes how we think about ``AI for productivity''. Different tools target different friction types: large language models may reduce search costs by summarising and organising information, but can also raise monitoring costs if their outputs require additional checking; automation can reduce routine monitoring burdens, but may increase adaptation costs if it creates brittle workflows; and decision-support can reduce decision costs only if it aligns with governance and accountability \citep{BMA2024AI,Raza2024Pathways}. The implication is that the optimal role for AI is not uniform across healthcare. It depends on the composition of transaction-cost frictions within a role or pathway. Without a way to measure that composition, health systems are left choosing tools based on plausibility, marketing, or local enthusiasm rather than system-level need.

In this study we operationalise a task-level TCE approach to healthcare work. Using occupational task statements and frequency weights, we classify tasks into the four transaction-cost categories and assign a bounded transaction-cost intensity score to each task. We then aggregate these task-level measures to the occupation level to estimate both (i) mean transaction-cost intensity and (ii) within-occupation heterogeneity. The resulting ``friction map'' is intended as a decision aid: it distinguishes roles where targeted task fixes are likely to suffice from those requiring structural redesign or mixed augmentation strategies. Finally, we test whether clinician and non-clinician occupations differ systematically in both the level and composition of transaction-cost frictions, thereby identifying where digital and AI tools are most likely to be strategically valuable.

\section{Methods}
\subsection{Data source}
We used the O*NET database of occupational task statements and task frequency/importance weights \citep{ONETDatabase}. Task statements were restricted to health-related occupations and deduplicated by task text to avoid redundant coding.

\subsection{LLM-based task coding and scoring}
\label{sec:llm_scoring}

Task statements are short natural-language descriptions of work activities and are not pre-labelled for transaction-cost constructs. We therefore treated the coding problem as a structured classification task: given a task statement, assign (i) a single dominant transaction-cost category and (ii) bounded ordinal scores capturing the coordination burden of executing that task once.

We used a large language model (LLM) accessed via an enterprise-hosted API (Azure OpenAI; deployment \texttt{gpt-4.1}; API version \texttt{2024-12-01-preview}). The LLM was used as a \emph{constrained scorer}: it was instructed to output only a machine-parseable record conforming to a fixed schema (below), rather than free text.

For each task statement, the model was provided with the task text together with clear definitions of the four transaction-cost categories used in this study (information search, decision and bargaining, monitoring and enforcement, and adaptation and coordination). The model was instructed to consider the task as it would typically be performed once and to return a structured record describing both the dominant source of coordination cost and the factors contributing to it. Specifically, the output captured:
\begin{itemize}
  \item the single dominant transaction-cost category associated with the task;
  \item an overall transaction-cost intensity score (0--5), summarising the coordination burden involved in performing the task once;
  \item a set of underlying driver scores (each 0--3) reflecting distinct sources of coordination cost, including uncertainty, ease of measurement, asset specificity, interdependence with other actors, and exposure to opportunism;
  \item where applicable, brief qualitative tags indicating the type of coordination friction most salient for the task (for example, information search, decision coordination, monitoring, or adaptation).
\end{itemize}

To minimise circularity, the model was shown the task statement text but not the occupation title during scoring.

All model outputs were validated in code against: (i) JSON parsability; (ii) categorical membership constraints; and (iii) numeric bounds for each ordinal field. Invalid outputs were not silently accepted. Instead, the system performed automated repair using a bounded retry loop: if the response failed parsing or violated constraints, the same task was re-submitted with an explicit error message describing the schema violation and requesting a corrected JSON object. If repeated attempts failed, the task was recorded with \texttt{error} metadata and excluded from downstream aggregation. Where present, a \texttt{repaired} flag indicates that a non-conforming response required at least one repair attempt before acceptance.

Because identical task texts recur across multiple occupations, task statements were deduplicated by exact string match. Each unique task text was scored once, then joined back to the full task-occupation table to recover occupation membership and frequency weights. This reduces cost and enforces consistency: identical task statements receive identical transaction-cost labels and scores wherever they appear.

The aim of the scoring step is to operationalise a theoretical construct (transaction-cost burden) over a large set of heterogeneous, natural-language task statements. LLMs are particularly suited to this style of constrained semantic classification when (i) the label space is clearly defined, (ii) outputs are structurally validated, and (iii) the downstream analysis depends on bounded ordinal scores rather than model-generated prose. This approach aligns with implementation-focused guidance emphasising structured evaluation and workflow relevance over model novelty \citep{Raza2024Pathways,BMA2024AI}.

\subsection{Aggregation to occupations}
For occupation $o$ with tasks $i$, and frequency weights $w_i$, the occupation-level transaction-cost index (TCI) was computed as:
\begin{equation}
\mathrm{TCI}_o = \frac{\sum_i w_i\, \mathrm{tc\_intensity}_i}{\sum_i w_i}.
\end{equation}
Within-occupation dispersion was computed as the square root of the corresponding weighted variance (TCI$_\mathrm{sd}$). Category shares $\mathrm{share}_{o,c}$ were computed as the fraction of total frequency weight in category $c$.

We treated occupations as the unit of inference to avoid inflated significance from large task counts. Clinician vs non-clinician differences in TCI, TCI$_\mathrm{sd}$, and category shares were tested with Mann-Whitney $U$ tests. Effect sizes were summarised using Cliff's $\delta$. Multiple testing was corrected for using the Benjamini-Hochberg procedure.

\section{Results}
\begin{table}[t]
\centering
\caption{Headline role-level summaries (frequency-weighted). Values are computed from task-level labels and intensities using O*NET task frequency weights.}
\label{tab:headline}
\begin{tabular}{lrr}
\hline
 & Non-clinician & Clinician\\
\hline
Weighted mean TCI & 2.786 & 3.352\\
SEARCH\_INFO share & 0.180 & 0.267\\
BARGAIN\_DECIDE share & 0.013 & 0.073\\
MONITOR\_ENFORCE share & 0.404 & 0.352\\
ADAPT\_COORDINATE share & 0.403 & 0.308\\
\hline
\end{tabular}
\end{table}

\begin{table}[t]
\centering
\caption{Occupation-level differences between clinician and non-clinician occupations. Mann--Whitney $U$ tests (two-sided) with Benjamini--Hochberg false discovery rate (FDR) correction; Cliff's $\delta$ reported as effect size. Positive $\delta$ indicates higher values for clinician occupations.}
\label{tab:occ_tests}
\begin{tabular}{lrrrrrr}
\hline
Variable & $p$ & $p_{\mathrm{FDR}}$ & $U$ & $\delta$ & $\tilde{x}_{\mathrm{clin}}$ & $\tilde{x}_{\mathrm{non}}$\\
\hline
TCI & $2\times10^{-6}$ & $1.5\times10^{-5}$ & 1248.0 & 0.727 & 3.498 & 2.793\\
share\_BARGAIN\_DECIDE & $9.66\times10^{-4}$ & $2.90\times10^{-3}$ & 1075.5 & 0.489 & 0.059 & 0.000\\
share\_SEARCH\_INFO & $6.60\times10^{-3}$ & $1.32\times10^{-2}$ & 1025.5 & 0.419 & 0.266 & 0.180\\
share\_MONITOR\_ENFORCE & $5.47\times10^{-2}$ & $6.56\times10^{-2}$ & 508.0 & -0.297 & 0.286 & 0.385\\
share\_ADAPT\_COORDINATE & $5.14\times10^{-2}$ & $6.56\times10^{-2}$ & 505.0 & -0.301 & 0.340 & 0.445\\
TCI\_sd & $5.41\times10^{-1}$ & $5.41\times10^{-1}$ & 791.0 & 0.095 & 0.658 & 0.628\\
\hline
\end{tabular}
\end{table}

\begin{figure}[t]
\centering
\includegraphics[width=0.9\linewidth]{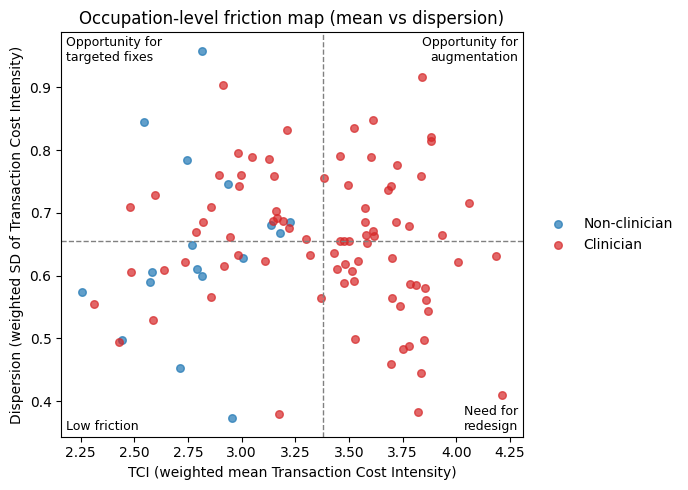}
\caption{Transaction‑cost friction across healthcare occupations.
Occupations are plotted by mean transaction‑cost intensity and within‑occupation dispersion across tasks, capturing coordination burden (search, decision, monitoring, adaptation) rather than task complexity; clinician and non‑clinician roles occupy distinct coordination regimes, implying uneven digital opportunity.}
\label{fig:friction}
\end{figure}

Figure~\ref{fig:friction} shows the ``friction map'': each point is an occupation positioned by mean friction (TCI) and dispersion (TCI$_\mathrm{sd}$). Quadrants separate occupations likely to benefit from targeted task-level fixes (low mean, high dispersion) from those requiring structural redesign (high mean, low dispersion) or mixed augmentation strategies (high mean, high dispersion).

\section{Discussion}
Much of the contemporary discourse on digital transformation in healthcare rests on an implicit orthodoxy: that the primary determinant of benefit is technological capability, and that gains will follow from applying increasingly powerful tools to increasingly granular tasks. This view underpins common practices such as ranking tasks by apparent opportunity for automation, focusing evaluation on accuracy or time saved, and assuming that adoption barriers are primarily cultural or behavioural. Our findings challenge this framing. They show that healthcare work is not a homogeneous substrate awaiting automation, but a set of roles embedded in markedly different coordination regimes. The dominant constraints on performance are frequently not the technical difficulty of tasks, but the structure and distribution of transaction costs surrounding them. Ignoring this structure risks systematic misinterpretation of both success and failure in digital adoption.

This study shows that healthcare work differs systematically in both the level and composition of transaction-cost burdens, and that these differences align with occupational role. Clinician occupations exhibit higher overall transaction-cost intensity than non-clinician occupations, alongside a greater share of costs arising from search and decision-related coordination, whereas non-clinician work is more heavily weighted toward monitoring, enforcement, and routine coordination. Importantly, these differences are not marginal: they reflect distinct coordination regimes rather than variation around a common mean.

The central implication is not that some roles are somehow more amenable to automation than others, but instead that different kinds of digital opportunity exist in different parts of the healthcare workforce. Treating digital adoption as a uniform productivity intervention obscures this heterogeneity and risks mis-targeting both technology and evaluation effort.

By making transaction-cost heterogeneity visible at scale, this study reframes digital opportunity as a structural property of work rather than an intrinsic property of technology. The implication is that uneven returns from digital adoption are not anomalous, nor primarily a function of poor implementation, but a predictable consequence of deploying similar interventions into fundamentally different coordination environments. This challenges the expectation that broad productivity gains should accrue uniformly across occupations, and instead suggests that ceilings on benefit are role-specific and shaped by institutional and organisational design. The contribution of this work is therefore not to identify which technologies should be adopted, but to demonstrate why opportunity is uneven in the first place—and why strategies grounded in uniform automation narratives are likely to disappoint.

These results suggest a more precise approach to ``AI for productivity'': interventions should target the dominant friction type in each role. For clinicians, tools that compress information search and decision coordination may yield more leverage than generic documentation automation. For non-clinician roles, reducing monitoring burdens and exception handling costs may dominate.

This study focuses on transaction costs—costs of coordination, information, and governance—rather than on direct production costs such as labour time, consumables, or capital intensity. Digital and AI interventions may act on these direct costs through mechanisms that are not captured here, including automation of physical tasks, substitution of labour, reduction in error-related reworking etc. However, the emphasis on transaction costs is deliberate and theoretically motivated. In healthcare failures of digital adoption commonly arise not because technologies fail to perform their intended function, but because they interact poorly with existing coordination structures—adding monitoring burden, shifting work across roles, or increasing exception-handling and governance overhead. Transaction costs therefore play a critical mediating role: they determine whether reductions in direct production costs translate into realised system-level benefit or are offset by new forms of coordination work. By focusing on transaction costs, this study addresses a dimension of digital value that is often implicit, rarely measured, and frequently invoked only post hoc to explain disappointing outcomes. The results should thus be interpreted as complementary to analyses of direct cost reduction, providing insight into why nominally effective digital interventions

It is also important to recognise that digital and AI interventions may not reduce transaction costs so much as redistribute them within the system, and that such redistribution may in some cases be appropriate. Coordination burdens can shift across roles, organisational levels, or stages of work—for example, from execution to oversight, from frontline clinicians to supervisory or support functions, or from routine processes to exception handling and governance. Given that clinician labour is both scarce and poorly scalable, reallocating transaction costs away from clinician time may represent a rational system-level trade-off, even if total coordination burden remains unchanged. Interventions that reduce information search or decision effort at the point of care may therefore increase monitoring, audit, documentation, or liability-related work elsewhere, particularly in safety-critical environments. These displacement effects are often invisible in local or task-level evaluations and may only become apparent at organisational or system scale. As a result, apparent efficiency gains for clinicians may coexist with stable or increased transaction costs overall, while still constituting a net improvement.

Taken together, these considerations strengthen the focus on transaction costs. If digital and AI interventions can both reduce and reallocate coordination burdens, then understanding the structure and distribution of transaction costs becomes essential to interpreting their net effect. Direct production efficiencies are realised locally and immediately, whereas transaction costs operate systemically: they shape how work is authorised, coordinated, monitored, and adapted as interventions scale. It is therefore transaction costs that determine whether improvements in technical efficiency persist, dissipate, or are neutralised by new forms of organisational work. By foregrounding transaction-cost structure, this study provides an explanatory lens for why similar digital interventions can generate durable benefit in some roles yet yield fragile or contested gains in others, even when their technical performance is comparable.

\subsection{Limitations}
O*NET reflects US occupational descriptions and may not map perfectly to other health systems. Task statements are abstractions and do not capture local workflow design. LLM-based coding introduces model and prompt dependence; we mitigate this with schema constraints and range validation but do not claim ground truth labels. Finally, frequency weights are proxies for relative task prominence, not time-on-task.

\section{Conclusions}
This study challenges the assumption that digital value in healthcare is determined primarily by technological capability, showing instead that opportunity is structured by how coordination costs are distributed across different kinds of work. Task-level transaction-cost coding offers a scalable way to quantify organisational frictions in healthcare work. The resulting friction map distinguishes redesign opportunities from augmentation opportunities and supports theory-grounded targeting of digital interventions.

\section*{Data availability}
O*NET data are publicly available from the O*NET Resource Center \citep{ONETDatabase}. The derived task-level labels and occupation-level aggregates can be regenerated from O*NET inputs using the provided code.

\section*{Code availability}
A reproducible analysis pipeline (LLM coding, aggregation, and summary-pack generation) is provided in the accompanying repository (to be supplied by the authors upon submission).

\section*{Ethics}
This study uses publicly available occupational descriptors and contains no patient-level data.

\section*{Competing interests}
The author declares no competing interests.

\bibliographystyle{unsrtnat}
\bibliography{references}

\end{document}